\documentclass[conference]{IEEEtran}
\IEEEoverridecommandlockouts

\usepackage{cite}
\usepackage{amsmath,amssymb,amsfonts}
\usepackage{algorithmic}
\usepackage{graphicx}
\usepackage{textcomp}
\usepackage{xcolor}
\usepackage{url}

\usepackage{tikz}
\usetikzlibrary{positioning,arrows.meta,fit,backgrounds,shapes.geometric}
\usepackage{pgfplots}
\pgfplotsset{compat=1.17}

\usepackage{booktabs}
\usepackage{multirow}
\usepackage{tabularx}

\usepackage[ruled,vlined]{algorithm2e}

\definecolor{gemblue}{RGB}{31,78,121}
\definecolor{gemsand}{RGB}{245,240,228}
\definecolor{gemobsidian}{RGB}{40,40,40}

\tikzset{
  block/.style={draw, fill=gemblue!8, rounded corners=3pt,
                align=center, font=\small, inner sep=4pt,
                minimum height=9mm},
  edgenode/.style={draw, fill=gemblue!20, rounded corners=3pt,
                   align=center, font=\small, inner sep=4pt,
                   minimum height=9mm},
  store/.style={draw, fill=gemsand, rounded corners=3pt,
                align=center, font=\small, inner sep=4pt,
                minimum height=9mm},
  flow/.style={-Stealth, thick, gemblue!70}
}

\def\BibTeX{{\rm B\kern-.05em{\sc i\kern-.025em b}\kern-.08em
    T\kern-.1667em\lower.7ex\hbox{E}\kern-.125emX}}

\begin{document}

\title{TimeLens: On-Device Artifact Recognition with\\
Retrieval-Augmented Question Answering\\
for the Grand Egyptian Museum}

\author{
\IEEEauthorblockN{1\textsuperscript{st} Rawan Hesham}
\IEEEauthorblockA{
  \textit{Faculty of Computers and AI}\\
  \textit{Capital University}\\
  Cairo, Egypt\\
  rawan\_20220167@fci.capu.edu.eg}
\and
\IEEEauthorblockN{2\textsuperscript{nd} Ali Ashraf}
\IEEEauthorblockA{
  \textit{Faculty of Computers and AI}\\
  \textit{Capital University}\\
  Cairo, Egypt\\
  AliEldin\_20220296@fci.capu.edu.eg}
\and
\IEEEauthorblockN{3\textsuperscript{rd} Amr Ahmed}
\IEEEauthorblockA{
  \textit{Faculty of Computers and AI}\\
  \textit{Capital University}\\
  Cairo, Egypt\\
  Amr\_20210632@fci.capu.edu.eg}
\and
\IEEEauthorblockN{4\textsuperscript{th} Malak Alaa}
\IEEEauthorblockA{
  \textit{Faculty of Computers and AI}\\
  \textit{Capital University}\\
  Cairo, Egypt\\
  Malak\_20220503@fci.capu.edu.eg}
\and
\IEEEauthorblockN{5\textsuperscript{th} Omar Ahmed}
\IEEEauthorblockA{
  \textit{Faculty of Computers and AI}\\
  \textit{Capital University}\\
  Cairo, Egypt\\
  Omar\_20220310@fci.capu.edu.eg}
\and
\IEEEauthorblockN{6\textsuperscript{th} Omar Wagih}
\IEEEauthorblockA{
  \textit{Faculty of Computers and AI}\\
  \textit{Capital University}\\
  Cairo, Egypt\\
  Omar\_20220324@fci.capu.edu.eg}
}

\maketitle

\begin{abstract}
TimeLens is an AI-powered bilingual mobile guide for the Grand Egyptian Museum (GEM).
Pointing a phone at an exhibit, a visitor sees the artifact recognized in real time
and can ask follow-up questions answered in English or Arabic. The work addresses
three problems specific to in-gallery deployment: fine-grained visual similarity among
51 catalogued artifacts (many near-identical Ramesside statues), the gap between
curated training data and handheld camera conditions, and the risk of an AI guide
stating unsupported historical facts. Two engineering contributions are reported.
First, an on-device artifact detector was developed through a data-quality-driven
iteration study --- from foundation-model auto-annotation (YOLO-World), through
spatial label-cleaning rules, to a fully hand-annotated dataset --- isolating label
quality as the decisive factor: the final YOLOv8n model resolves every previously
failing class while remaining a 5.97 MB TensorFlow Lite asset that runs in real time
on a mid-range phone (mAP@0.5 = 0.995, mAP@0.5:0.95 = 0.924). Second, a bilingual
Retrieval-Augmented Generation (RAG) guide, grounded in a 108-record ChromaDB
knowledge base, was benchmarked across seven candidate language models, with Gemma 4
E2B (Q4\_K\_M) selected; ten targeted optimizations reduce end-to-end latency from
over 30 s to approximately 10 s. Both subsystems are integrated in a production
Flutter application with bilingual interface, museum location gating, and
text-to-speech support.
\end{abstract}


\section{Introduction}

Fine-grained artifact recognition in museum environments is difficult due to
background clutter, low-light capture, visual overlap between classes, and
open-set uncertainty. At the Grand Egyptian Museum (GEM) the challenge is acute:
the target collection contains 51 catalogued artifacts, many of which are
near-identical Ramesside statues differing only in subtle pose, headdress, or
accompanying standards. A tourist standing in front of two visually similar statues
does not care about model internals; they care whether the app gives a trustworthy
answer quickly and in their own language.

Current museum-technology offerings fail on at least one of three requirements for
GEM deployment: real-time on-device recognition (cloud APIs introduce latency and
connectivity dependence), Arabic-language support (most systems target Western
audiences), and hallucination-free answers (ungrounded language models routinely
invent plausible-sounding historical facts) \cite{abdelhady2025enhancing}. The global
Augmented Reality market exceeded USD 50 billion in 2023 and is projected to surpass
USD 110 billion by 2027 \cite{PwCXR2023}; Egypt welcomed approximately 14.9 million
international visitors in 2023 \cite{UNWTO2024}, and over 70\% of tourists rely on
digital tools during visits \cite{TGMEgypt2025}, confirming strong demand for a
solution that satisfies all three requirements simultaneously.

\textbf{TimeLens} addresses this gap through two engineering contributions. First, a
compact YOLOv8n detector \cite{yolov8} is trained on a carefully curated,
video-derived dataset of GEM artifacts and exported to TensorFlow Lite with
end-to-end non-maximum suppression, enabling real-time on-device recognition with no
per-frame network round-trip. Second, a bilingual Retrieval-Augmented Generation
(RAG) guide \cite{lewis2020rag} grounds every answer in a curated 108-record
ChromaDB knowledge base, preventing fabrication and supporting both English and
Arabic. Both subsystems are integrated in a production Flutter application.

The key findings of this paper are: (i) label quality dominates model architecture
and input resolution for fine-grained museum artifact detection; (ii) a fully local
bilingual RAG pipeline achieves hallucination-free answers at interactive latency;
and (iii) the combined system runs on commodity phones and a single small GPU, making
it suitable for real museum deployment.

\section{Related Work}

\subsection{On-Device Object Detection for Cultural Heritage}

Single-stage detectors of the YOLO family provide a strong speed/accuracy balance
for mobile deployment and export cleanly to TensorFlow Lite for on-device inference
\cite{yolov8}. Comparative studies of museum-exhibit recognition from video-derived
datasets report that YOLOv8 reaches very high mean Average Precision while remaining
light enough for real-time mobile use, outperforming heavier backbones such as VGG-16
and ResNet on the speed/accuracy trade-off \cite{ipalakova2025comparative}. Adaptive
CNN approaches have likewise been used to enrich museum interaction through automatic
artifact recognition \cite{wen2024enhancing}. EfficientNet-based systems have been
applied to cultural-landmark recognition in smart tourism with strong
results \cite{11239039}.

\subsection{Foundation-Model Auto-Annotation}

Manually annotating tens of thousands of video frames is prohibitive for a
student-led project, motivating auto-annotation with open-vocabulary,
text-prompted detectors such as YOLO-World \cite{yoloworld}. Such foundation models
handle generic concepts well but degrade on fine-grained, domain-specific
objects --- precisely the near-identical statues that dominate the GEM collection ---
producing systematically mislocalized or sibling-confused labels. This limitation is
the central motivation for our data-quality study.

\subsection{Retrieval-Augmented Generation}

Retrieval-Augmented Generation (RAG) grounds a language model's output in passages
retrieved from a knowledge base rather than its parametric memory, reducing
fabrication in knowledge-intensive tasks \cite{lewis2020rag}. Recent work brings
context-aware LLM assistants into AR settings \cite{qorbani2025teaching}. Quantized
models run locally through runtimes such as Ollama \cite{ollama}, while multilingual
sentence embeddings make a single bilingual retrieval index practical. Robustness
challenges specific to heritage capture --- low contrast, reduced lighting, and crowd
occlusion --- are active research topics \cite{liu2025yolov8,dai2024occlusion}.

\subsection{Research Gaps}

Prior work leaves three combined gaps that TimeLens addresses: (i) label quality for
fine-grained, domain-specific detection; (ii) honest evaluation of video-frame
datasets where random splits leak near-duplicate frames; and (iii) grounded, bilingual
generation for public-facing museum guides.

\section{System Architecture}

TimeLens separates concerns across two principal layers with a clear boundary. On the
phone, an \emph{on-device perception layer} turns each camera frame into a tracked,
classified artifact. A \emph{knowledge layer} then grounds any follow-up question in
curated museum data before the language model phrases the answer.

\subsection{High-Level Block Diagram}

\begin{figure}[!t]
\centering
\resizebox{\columnwidth}{!}{%
\begin{tikzpicture}[
    node distance=5mm and 8mm,
    block/.style={
        draw,
        rectangle,
        rounded corners=1pt,
        align=center,
        minimum height=8mm,
        font=\footnotesize
    },
    store/.style={
        draw,
        rectangle,
        rounded corners=1pt,
        align=center,
        minimum height=8mm,
        font=\footnotesize
    },
    flow/.style={->,>=latex,thick}
]

\node[block,text width=20mm] (app)
{Mobile App\\(Flutter)\\{\scriptsize YOLOv8n TFLite}};

\node[block,right=of app,text width=18mm] (api)
{FastAPI\\RAG Service};

\node[block,right=of api,text width=18mm] (gemma)
{Ollama\\Gemma 4 E2B};

\node[store,below=6mm of api,text width=18mm] (chroma)
{ChromaDB\\{\scriptsize EN/AR collections}};

\draw[flow] (app) -- node[above,font=\tiny]{HTTPS} (api);
\draw[flow] (api) -- node[below,font=\tiny]{answer} (app);

\draw[flow] (api) -- node[above,font=\tiny]{prompt} (gemma);

\draw[flow] (api) -- node[right,font=\tiny]{retrieve} (chroma);

\end{tikzpicture}
}
\caption{System block diagram. Detection is fully on-device; only chat questions reach the FastAPI RAG service.}
\label{fig:block}
\end{figure}
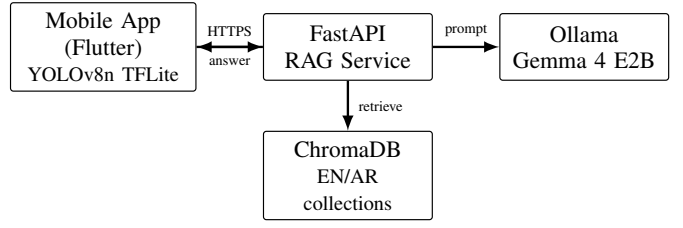

Fig.~\ref{fig:block} shows the runtime data flow. Detection runs entirely on the
phone; only a chat question crosses the network, reaching the FastAPI service that
retrieves from ChromaDB and prompts the locally served Gemma model.

\subsection{Scan-to-Answer Sequence}

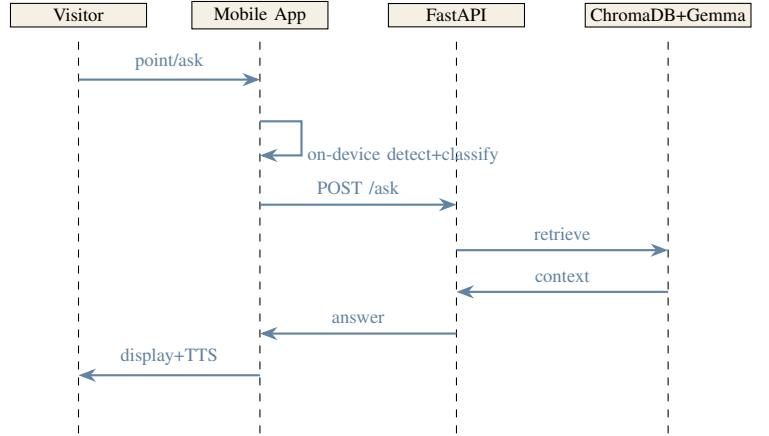
\begin{figure}[htbp]
\centering
\begin{tikzpicture}[font=\scriptsize, node distance=0mm]
  \def\ytop{0} \def\ybot{-5.6}
  \foreach \x/\lbl in {0/Visitor, 2.4/{Mobile App}, 5.0/FastAPI, 7.8/{ChromaDB+Gemma}}{
    \node[draw, fill=gemsand, minimum width=18mm, font=\scriptsize,
          inner sep=2pt, align=center] at (\x,\ytop) {\lbl};
    \draw[dashed] (\x,\ytop-0.35) -- (\x,\ybot);
  }
  \draw[flow] (0,-0.85)  -- node[above]{\scriptsize point/ask} (2.4,-0.85);
  \draw[flow] (2.4,-1.4) -- ++(0.55,0) |- (2.4,-1.85)
    node[right, xshift=5mm]{\scriptsize on-device detect+classify};
  \draw[flow] (2.4,-2.5) -- node[above]{\scriptsize POST /ask} (5.0,-2.5);
  \draw[flow] (5.0,-3.1) -- node[above]{\scriptsize retrieve} (7.8,-3.1);
  \draw[flow] (7.8,-3.65)-- node[above]{\scriptsize context} (5.0,-3.65);
  \draw[flow] (5.0,-4.2) -- node[above]{\scriptsize answer} (2.4,-4.2);
  \draw[flow] (2.4,-4.75)-- node[above]{\scriptsize display+TTS} (0,-4.75);
\end{tikzpicture}
\caption{Scan-to-answer sequence. Identity is established on-device; the backend
retrieves context and phrases the grounded answer.}
\label{fig:sequence}
\end{figure}

Fig.~\ref{fig:sequence} traces one interaction from end to end. Identity is decided
on-device the moment the visitor frames an artifact; the backend is contacted only
when a question is asked.

\subsection{On-Device Inference Pipeline}

Each camera frame arrives as sensor-rotated YUV420 data, converted to RGB, rotated
$90^{\circ}$ to match the Android sensor orientation, resized to $640\!\times\!640$,
and normalized before inference. The YOLOv8n model is exported with NMS baked in, so
a single forward pass returns a fixed-size \texttt{[1,300,6]} tensor of
\texttt{(xyxy, score, class)} rows --- removing any post-processing decode on the
device. A confidence filter (global threshold 0.75, raised to 0.90 for the
visually near-identical Statue of Siptah) then keeps only trustworthy detections.

\subsection{Technology Stack}

\begin{itemize}
\item \textbf{Mobile app:} Flutter with Provider state management.
\item \textbf{On-device detection:} YOLOv8n (Ultralytics) exported to TensorFlow
      Lite (FP16) with end-to-end NMS.
\item \textbf{RAG backend:} FastAPI + LlamaIndex + LangChain; Ollama serving
      Gemma 4 E2B (Q4\_K\_M); ChromaDB vector store; multilingual MiniLM embeddings.
\item \textbf{Training:} Ultralytics 8.4.45 / PyTorch; YOLO-World for
      auto-annotation; Roboflow for hand annotation; NVIDIA GTX 1650 Ti (4 GB).
\item \textbf{Serving:} RunPod NVIDIA RTX A5000 (24 GB VRAM) via ngrok tunnel.
\end{itemize}

\section{On-Device Artifact Detection}

\subsection{Dataset and Constraints}

The source material consists of approximately 50 in-gallery videos of GEM artifacts
captured before public opening, yielding roughly 30,000 candidate frames across 51
classes at about one frame per second. Deployment constraints rule out cloud
detectors: inference must run on-device with no per-frame round-trip, in real time
($\geq\!2.5$ FPS on mid-tier phones), with a compact model ($<\!8$ MB TFLite asset),
and be robust to handheld capture. These constraints led to YOLOv8-nano with FP16
TensorFlow Lite deployment \cite{yolov8,ipalakova2025comparative}.

\subsection{Label-Quality Iteration Study}

Manual annotation of 30k images is infeasible for a student team, motivating an
auto-annotation-first pipeline followed by progressive quality improvement.

\textbf{v1 --- Auto-Annotated Baseline.}
Labels were produced by YOLO-World (\texttt{yolov8s-worldv2}) \cite{yoloworld} using
per-class text prompts. A YOLOv8n model trained on these labels at $320\!\times\!320$
reached mAP@0.5 $=0.751$, but per-class analysis revealed ten classes below
mAP@0.5 $=0.30$. Two failure modes appeared: Mode~A (nine classes), where YOLO-World
boxed the wrong object (a wall card, a bench, a neighboring statue); and Mode~B (one
class), sibling-class confusion between a bust and full-body statues. The architecture
was capable; the labels were the bottleneck.

\textbf{v2 --- Spatial Sanity Rules.}
Four geometric rules were applied to the YOLO-World output:
(1) restrict the ``stela'' prompt to the two true stela classes;
(2) drop boxes whose center lies in the bottom quarter of the frame ($c_y > 0.75$),
which are almost always floor signs or bystanders;
(3) pick the largest-area box rather than the highest-confidence one;
(4) drop boxes whose center deviates from the per-video median by more than 0.20.
These rules dropped 6.9\% of frames (leaving 27,134), raised mAP@0.5 to 0.867
(+0.116 from a labeling change alone), and reduced broken classes from ten to three.

\textbf{v2@640 --- Resolution and End-to-End NMS.}
Raising input to $640\!\times\!640$ and exporting with end-to-end NMS left mAP@0.5
unchanged at 0.867 but lifted mAP@0.5:0.95 from 0.586 to 0.777 (+0.191): higher
resolution improves localization but cannot fix classes whose labels are wrong. The
data-quality ceiling was unmistakable.

\textbf{v3 --- Full Hand Annotation.}
All 51 classes were re-annotated by hand in Roboflow (24,384 images, 70/20/10 split),
drawing full-artifact boxes, avoiding peripheral objects, and carefully distinguishing
similar artifacts. Warm-started from v2@640 weights and trained for 100 epochs on a
GTX 1650 Ti, v3 reached precision 0.998, recall 1.000, mAP@0.5 0.995, and
mAP@0.5:0.95 0.924, with zero broken classes. The validation classification loss
collapsed from 0.892 to 0.166.

\subsection{Auto-Annotation Algorithm}

Algorithm~\ref{alg:autoannotate} formalizes the v1$\rightarrow$v2 spatial-cleaning
step.

\begin{algorithm}[htbp]
\small
\SetAlgoLined
\KwIn{video frames $F$; per-class text prompts $P$; true-stela class set $S$}
\KwOut{one labeled box per usable frame}
\ForEach{frame $f \in F$ (class $c$ from its folder)}{
  $B \leftarrow \textsc{YoloWorld}(f, P)$\;
  \If{$c \notin S$}{remove stela-prompt detections from $B$}
  remove boxes with center $c_y > 0.75$\;
  \If{$B \neq \emptyset$}{
    $b^\star \leftarrow \arg\max_{b\in B}\;\mathrm{area}(b)$\;
    drop $b^\star$ if its center deviates $>0.20$ from per-video median\;
    \If{$b^\star$ kept}{emit $(b^\star, c)$}
  }
}
\caption{Auto-annotation with spatial sanity rules (v1$\rightarrow$v2)}
\label{alg:autoannotate}
\end{algorithm}

\subsection{Detection Results}

\begin{figure}[htbp]
\centering
\begin{tikzpicture}
\begin{axis}[
  ybar, symbolic x coords={v1,v2@320,v2@640,v3},
  xtick=data, ymin=0, ymax=1.12,
  ylabel={mAP@0.5}, bar width=13pt,
  width=0.9\columnwidth, height=5.2cm,
  nodes near coords, every node near coord/.append style={font=\scriptsize},
  enlarge x limits=0.20,
  tick label style={font=\small},
  label style={font=\small}]
\addplot[fill=gemblue!55] coordinates
  {(v1,0.751)(v2@320,0.867)(v2@640,0.867)(v3,0.995)};
\end{axis}
\end{tikzpicture}
\caption{Detection mAP@0.5 across iterations. Label cleaning
(v1$\rightarrow$v2) and hand annotation (v2@640$\rightarrow$v3) drive the gains;
resolution scaling alone leaves mAP@0.5 flat.}
\label{fig:iteration}
\end{figure}
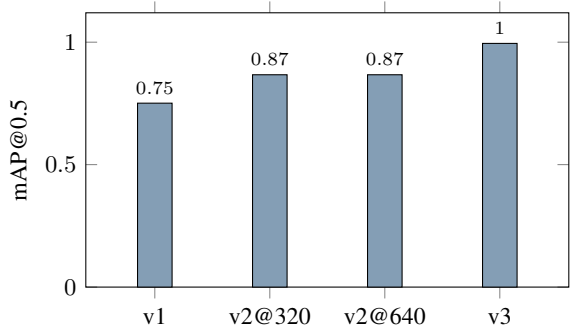

Table~\ref{tab:iteration} summarizes the four iterations.
Fig.~\ref{fig:iteration} plots the mAP@0.5 progression.
The dominant lesson is that label cleaning (+0.116 mAP@0.5, no model change) and
hand annotation (+0.128 mAP@0.5, +0.147 mAP@0.5:0.95) broke ceilings that
resolution scaling could not.

\begin{table}[htbp]
\centering
\caption{Detector Iteration Comparison (Validation Set)}
\label{tab:iteration}
\setlength{\tabcolsep}{4pt}
\small
\begin{tabular}{@{}lllcc@{}}
\toprule
\textbf{Iter.} & \textbf{Labels} & \textbf{imgsz} &
  \textbf{mAP$_{50}$} & \textbf{mAP$_{50:95}$}\\
\midrule
v1     & Auto (YOLO-World)      & 320 & 0.751 & ---  \\
v2     & Auto + 4 rules         & 320 & 0.867 & 0.586\\
v2@640 & Auto + rules           & 640 & 0.867 & 0.777\\
\textbf{v3} & \textbf{Hand-annotated} & \textbf{640}
       & \textbf{0.995} & \textbf{0.924}\\
\bottomrule
\end{tabular}
\end{table}

Table~\ref{tab:rescue} shows six representative classes rescued by hand annotation;
zero classes remain broken in v3.

\begin{table}[htbp]
\centering
\caption{Per-Class Rescue: Auto-Labels (v2) vs.\ Hand-Labels (v3), mAP@0.5:0.95}
\label{tab:rescue}
\small
\begin{tabular}{@{}lcc@{}}
\toprule
\textbf{Class} & \textbf{v2} & \textbf{v3}\\
\midrule
Double Statue of Ramesses II       & 0.000 & 0.995\\
Overseer Amenemhat                 & 0.000 & 0.940\\
Thutmose III Statue                & 0.000 & 0.941\\
Naktmin and Tiy                    & 0.269 & 0.977\\
Statue of King Akhenaten           & 0.281 & 0.955\\
Seated Statue of Thutmose III      & 0.440 & 0.862\\
\bottomrule
\end{tabular}
\end{table}

\subsection{Deployment Performance}

The deployed asset is FP16 TFLite at \textbf{5.97 MB}, running in approximately
800 ms per frame on a mid-tier Android phone (about 1.25 FPS). Although this falls
below the initial 2.5 FPS target, inference was executed asynchronously on a
separate thread, keeping the camera preview and UI responsive and non-blocking.
During field testing no misclassifications were observed across all 51 artifact
classes, with detection confidence consistently ranging between 75\% and 95\%.
Clean labels produce sharp confidence distributions: real artifacts cluster high
while off-target scenes fall well below the rejection threshold, enabling the raised
per-class thresholds (global 0.75; 0.90 for the false-positive-prone Siptah Stela).

\section{Bilingual RAG Museum Guide}

\subsection{Grounding Principles}

A public museum guide must not invent history. Three grounding principles govern the
design: (i) answers are constrained to retrieved museum context; (ii) retrieval is
language-matched so an Arabic question receives Arabic context; (iii) when relevant
facts are absent, the model is instructed to say so rather than fabricate. Running
inference locally removes per-request cost, rate limits, and visitor-query leakage.

\subsection{Knowledge Base}

The knowledge base comprises 108 records covering 54 unique artifacts, each documented
in English and Arabic. A preprocessing step fixes data-quality issues --- notably a
location field that had absorbed a material descriptor --- and preserves the original.
Each record is split into up to three semantically distinct chunks: an \emph{identity}
chunk (name, material, location, period), a \emph{description} chunk (appearance and
iconography), and a \emph{significance} chunk (historical and cultural context).
The 108 records yield \textbf{215 chunks} (109 English, 106 Arabic). English and
Arabic chunks are stored in \emph{separate} ChromaDB collections (\texttt{gem\_en} and
\texttt{gem\_ar}); a question is routed only to the matching collection.

\subsection{Retrieval Strategy}

Every request begins with language detection: more than 30\% Arabic-script characters
triggers Arabic routing, otherwise English. In \emph{camera mode}, the English title
of the detected artifact triggers a deterministic metadata lookup returning exactly
that artifact's chunks. In \emph{text mode}, a keyword-based query classifier assigns
the question to one of five categories (Table~\ref{tab:querycat}).

\begin{table}[htbp]
\centering
\caption{Text-Mode Query Categories and Retrieval Strategy}
\label{tab:querycat}
\small
\begin{tabularx}{\columnwidth}{@{}lX@{}}
\toprule
\textbf{Category} & \textbf{Retrieval strategy}\\
\midrule
Location & Metadata filter on \texttt{was\_found\_at} field (all matching chunks).\\
Period   & Metadata filter on historical-overview field.\\
Material & Metadata filter on material field.\\
Royal    & Similarity search ($k=8$) against language-matched collection.\\
Generic  & Similarity search ($k=8$) against language-matched collection.\\
\bottomrule
\end{tabularx}
\end{table}

\subsection{Model Selection}

Seven candidate models were benchmarked against 20 standardized bilingual questions.
Table~\ref{tab:funnel} summarizes the outcomes. \textbf{Gemma 4 E2B (Q4\_K\_M)} was
selected: it was the only model producing clean, non-hallucinated answers in
\emph{both} English and Arabic at the lowest cold-start VRAM.

\begin{table}[htbp]
\centering
\caption{Candidate Language Models and Benchmarking Outcome}
\label{tab:funnel}
\small
\setlength{\tabcolsep}{4pt}
\begin{tabular}{@{}llcc@{}}
\toprule
\textbf{Model} & \textbf{Quant.} & \textbf{VRAM (MB)} & \textbf{Verdict}\\
\midrule
Gemma 3n E2B     & default    & 1349  & Eliminated (hallucination)\\
Gemma 3n E4B     & Q3\_K\_M   & 2145  & Finalist\\
Gemma 4 E2B      & default    & 2021  & Eliminated (truncation)\\
\textbf{Gemma 4 E2B} & \textbf{Q4\_K\_M} & \textbf{1389--3585} & \textbf{Selected}\\
Llama 3.2 3B     & default    & 2565  & Finalist (Arabic unusable)\\
Qwen 3 4B        & default    & 2907  & Eliminated (thinking-mode)\\
Qwen 3.5 4B      & default    & 3485  & Eliminated (latency/VRAM)\\
\bottomrule
\end{tabular}
\end{table}

\subsection{Pipeline Optimizations}

\begin{figure}[htbp]
\centering
\begin{tikzpicture}
\begin{axis}[
  symbolic x coords={Baseline,{Metadata Fix},{All Opt},{Warm Cache}},
  xtick=data, ymin=0, ymax=38,
  ylabel={Avg.\ response time (s)},
  width=0.9\columnwidth, height=5.2cm,
  nodes near coords, point meta=explicit symbolic,
  enlarge x limits=0.18,
  tick label style={font=\small},
  label style={font=\small},
  x tick label style={rotate=12, anchor=east, font=\small}]
\addplot+[mark=*, thick, gemblue!80] coordinates {
  (Baseline,31)[$>$30]
  ({Metadata Fix},30.2)[30.2]
  ({All Opt},22.7)[22.7]
  ({Warm Cache},10.2)[10.2]};
\end{axis}
\end{tikzpicture}
\caption{End-to-end RAG latency across optimization phases (average over benchmark
questions): from over 30 s to a production-ready $\sim$10.2 s.}
\label{fig:latency}
\end{figure}
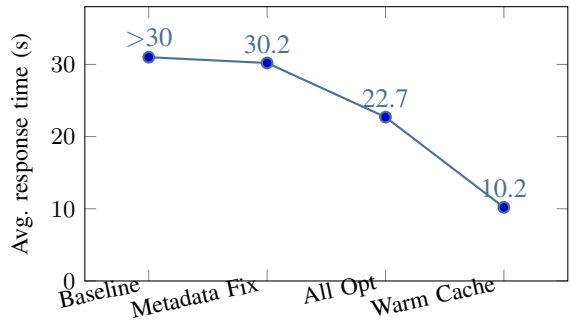

Ten targeted optimizations reduced end-to-end latency from over 30 s to approximately
10 s while improving answer quality. Key changes included: hierarchical
chunking (replacing flat documents with semantically distinct chunks), language-separated
collections, the query classifier with metadata filtering, mode-specific token
limits (\texttt{num\_predict} 750--1100 camera / 900--1200 text), Flash Attention
(\textasciitilde10--15\% faster inference), \texttt{keep\_alive}$=-1$ (permanent GPU
residency), a raised context window (1024$\rightarrow$4096 tokens),
\texttt{repeat\_penalty} raised from 1.1 to 1.3, \texttt{top\_p} = 0.9, and a
complete 15-royal Arabic name reference in the system prompt to eliminate pharaoh
name confusions. Fig.~\ref{fig:latency} plots the latency progression.

\subsection{RAG Evaluation Results}

On a 30-question evaluation set spanning text mode, camera mode, and out-of-dataset
queries in both languages, the pipeline averaged approximately 5.9 s per answer and
rated correct on all 30 responses, with no hallucinations, no truncations, and no
retrieval misses. All eight English text-mode, six Arabic text-mode, six English
camera-mode, three Arabic camera-mode, and three vague camera-mode questions returned
complete and accurate grounded answers. All four out-of-dataset queries (Rosetta
Stone, Old Kingdom artifacts, Great Pyramid construction, and the Rosetta Stone in
Arabic) were correctly refused without fabrication. Arabic answers averaged roughly
1 s slower than English, consistent with Arabic tokenizing to approximately twice the
length of equivalent English text.

\section{Market Context and Positioning}

Several tools operate in the museum-technology market, yet none fully meets GEM's
requirements. Table~\ref{tab:competitors} summarizes the competitive landscape.

\begin{table}[htbp]
\centering
\caption{Competitor Landscape for Museum Guide Technology}
\label{tab:competitors}
\small
\begin{tabularx}{\columnwidth}{@{}lX@{}}
\toprule
\textbf{Solution} & \textbf{Gap relative to GEM requirements}\\
\midrule
Google Arts \& Culture & No real-time camera recognition; no offline use\\
Smartify               & Requires connectivity; not tuned for Egyptian artifacts\\
QR-code systems        & Manual scanning interrupts visitor flow; no AI narration\\
Audio guides           & No interactivity, personalization, or AR overlay\\
\bottomrule
\end{tabularx}
\end{table}

The recurring gaps are: no real-time AI artifact recognition, limited or no offline
functionality, weak Arabic support, and no GEM-specific
experience \cite{cho2025augmenting,choiimmersive}. TimeLens fills these gaps with an
offline on-device detector (5.97 MB TFLite), a bilingual grounded RAG guide, and a
GEM-specific knowledge base --- positioning itself between heavy cloud-AI services and
simplistic QR-content systems \cite{ipalakova2025comparative}.

\section{Discussion}

\textbf{Label quality is first-order.} The controlled iteration study isolates label
quality as the decisive factor for fine-grained museum artifact detection: a labeling
change alone (+0.116 mAP@0.5) outweighs resolution scaling (0.000 mAP@0.5 gain).
Every previously failing class was rescued by hand annotation, confirming that
architectural choices are second-order once the label ceiling is removed.

\textbf{Grounding, not model size, prevents hallucination.} The decisive RAG gains
came from hierarchical chunking, language-separated collections, and query
classification --- not from a larger model. Gemma 4 E2B (Q4\_K\_M) at 1389--3585 MB
VRAM delivers hallucination-free bilingual answers where a 2565 MB Llama 3.2 3B
produces garbled Arabic.

\textbf{Limitations.} The knowledge base is closed: questions outside 108 records
receive limited answers. The query classifier is keyword-based and can misroute
questions that avoid expected vocabulary. The model has no memory across requests.
On the detection side, validation metrics reflect an image-level split; a
video-level (chronological) split would give a more conservative generalization
estimate.

\section{Conclusion}

This paper presented TimeLens, an AI-powered bilingual mobile guide for the Grand
Egyptian Museum comprising two integrated subsystems. The on-device detector
demonstrates that, for fine-grained museum artifacts, label quality is the primary
determinant of recognition performance: a compact YOLOv8n model trained on
hand-verified labels achieves mAP@0.5 = 0.995 and mAP@0.5:0.95 = 0.924 with zero
broken classes, fitting in a 5.97 MB TFLite asset. The bilingual RAG guide --- ChromaDB
retrieval grounding a quantized Gemma 4 E2B model behind FastAPI --- delivers factual
English/Arabic answers at approximately 10 s end-to-end latency with no hallucinations
across a 30-question evaluation. Both subsystems run on commodity hardware (a
mid-range phone and a single GPU) and are integrated in a production Flutter
application with bilingual interface and text-to-speech support.

Future work includes expanding the knowledge base and artifact dataset toward full-GEM
coverage, replacing the keyword query classifier with a learned intent model, adding
conversational memory, and conducting formal System Usability Scale studies in live
museum conditions.


\end{document}